\documentclass[letterpaper, 10 pt, journal, twoside]{IEEEtran}
\usepackage{times}
\usepackage[pdftex]{graphicx}
\usepackage{amsmath,amssymb,amsopn,amstext,amsfonts}
\usepackage{cancel}
\usepackage[space]{cite}
\usepackage{pdfsync}
\usepackage{balance}
\usepackage{color}
\usepackage{mathtools}
\usepackage{algpseudocode}
\usepackage{algorithm} 
\usepackage{bm}

\usepackage{float}
\usepackage{epstopdf}
\usepackage{pifont}
\usepackage{fixltx2e}
\usepackage{amsmath}
\usepackage{multirow}
\usepackage{url}
\usepackage{verbatim}
\usepackage{siunitx} 
\usepackage{epstopdf}

\usepackage{diagbox}
\usepackage{float}
\usepackage{epstopdf}
\usepackage{pifont}
\usepackage{fixltx2e}
\usepackage{multirow}
\usepackage{url}
\usepackage{verbatim}
\usepackage{siunitx} 
\usepackage{epstopdf}
\usepackage{bbding}

\usepackage{cancel}
\usepackage{pdfsync}
\usepackage{balance}
\usepackage{color}
\usepackage{mathtools}

\usepackage[linkcolor=black,citecolor=black,urlcolor=black,colorlinks=true]{hyperref}
\usepackage{booktabs}
\usepackage{makecell}
\usepackage{enumerate}
\usepackage{arydshln}
\usepackage[dvipsnames]{xcolor}

\title{H$_2$-Mapping: Real-time Dense Mapping Using Hierarchical Hybrid Representation}

\author{Chenxing Jiang$^{2,*}$, Hanwen Zhang$^{3,*}$, Peize Liu$^{2}$, Zehuan Yu$^{2}$, Hui Cheng$^{3}$, Boyu Zhou$^{1,\dag}$, Shaojie Shen$^{2}$
	\thanks{Manuscript received: June, 5, 2022; Accepted: August, 22, 2023. This paper was recommended for publication by Editor Javier Civera upon evaluation of the Associate Editor and Reviewers’ comments.}
	\thanks{\textsuperscript{*} \textbf{Equal contribution}. \textsuperscript{\dag} \textbf{Corresponding Author}. }
	\thanks{\textsuperscript{1} School of Artificial Intelligence, Sun Yat-Sen University, Zhuhai, China.}
	\thanks{\textsuperscript{2} Department of Electronic and Computer Engineering, The Hong Kong University of Science and Technology, Hong Kong, China.}
	\thanks{\textsuperscript{3} School of Computer Science and Engineering, Sun Yat-Sen University, Guangzhou, China}
	\thanks{\scriptsize \{\href{mailto:cjiangan@connect.ust.hk}{cjiangan@connect.ust.hk} \{\href{mailto:zhanghw66@mail2.sysu.edu.cn}{zhanghw66@mail2},\href{mailto:zhouby23@mail.sysu.edu.cn}{zhouby23@mail}\}.sysu.edu.cn\}}
	\thanks{Digital Object Identifier (DOI): see top of this page.}
}

\begin{document}
	
	\maketitle
	
	\begin{abstract}   
		Constructing a high-quality dense map in real-time is essential for robotics, AR/VR, and digital twins applications. 
		As Neural Radiance Field (NeRF) greatly improves the mapping performance, in this paper, we propose a NeRF-based mapping method that enables higher-quality reconstruction and real-time capability even on edge computers.
		Specifically, we propose a novel hierarchical hybrid representation that leverages implicit multiresolution hash encoding aided by explicit octree SDF priors, describing the scene at different levels of detail. This representation allows for fast scene geometry initialization and makes scene geometry easier to learn.
		Besides, we present a coverage-maximizing keyframe selection strategy to address the forgetting issue and enhance mapping quality, particularly in marginal areas. 
		To the best of our knowledge, our method is the first to achieve high-quality NeRF-based mapping on edge computers of handheld devices and quadrotors in real-time.
		Experiments demonstrate that our method outperforms existing NeRF-based mapping methods in geometry accuracy, texture realism, and time consumption. The code will be released at \url{https://github.com/SYSU-STAR/H2-Mapping}.
	\end{abstract}
	\begin{IEEEkeywords}
		Mapping; RGB-D Perception; Visual Learning
	\end{IEEEkeywords}
	
	 \IEEEpeerreviewmaketitle
	\setlength{\parskip}{0.05cm}
	\section{Introduction}
	% \subsection{jcx version}
	\IEEEPARstart{U}{sing} robots to build highly-detailed dense maps in real-time benefits advanced robot autonomous navigation, AR/VR, and digital twins applications. These maps enable robots to perform high-level tasks and provide humans with real-time feedback on the environment, allowing them to adjust the robot's tasks promptly as needed. Besides, high-fidelity maps serve as critical assets for AR/VR and digital twins. The automatic and faithful recreation of environments in real-time using robots can be more efficient and time-saving than manual or offline reconstruction methods.
	
	To be suitable for real-time and high-quality robot mapping in unknown environments with limited onboard computation power, a mapping system must meet four key requirements: (1) Adaptability to growing scenes, allowing the robot to dynamically expand the map without prior knowledge of the scene; (2) High level of detail; (3) Real-time capability and high memory efficiency; and (4) Novel view synthesis ability, which allows rendering high-quality images from views apart from the sparse input views. This is particularly important for creating scenes for AR/VR applications.
	\begin{figure}[t!]
		\centering
		\includegraphics[height=0.88\linewidth,width=1\linewidth]{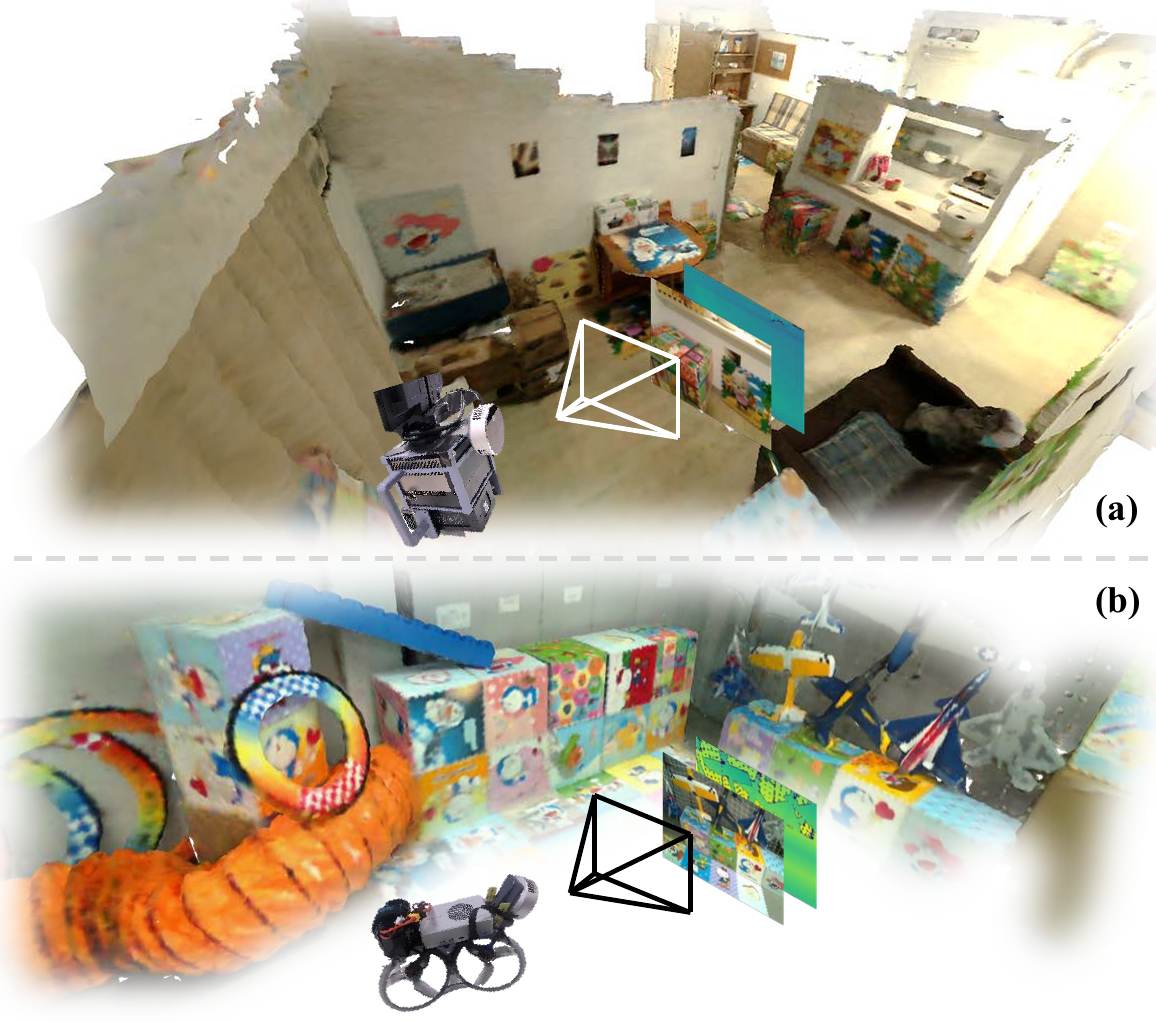}
		\caption{We tested our methods on a handheld device (a) and quadrotors (b). Our method builds a high-quality map in real-time on edge computers and can support robotic applications.}
		\label{fig:real_world_exp}
		\vspace{-1.6cm}
	\end{figure}
	\begin{figure*}[t!]
		\centering
		\includegraphics[width=1\textwidth]{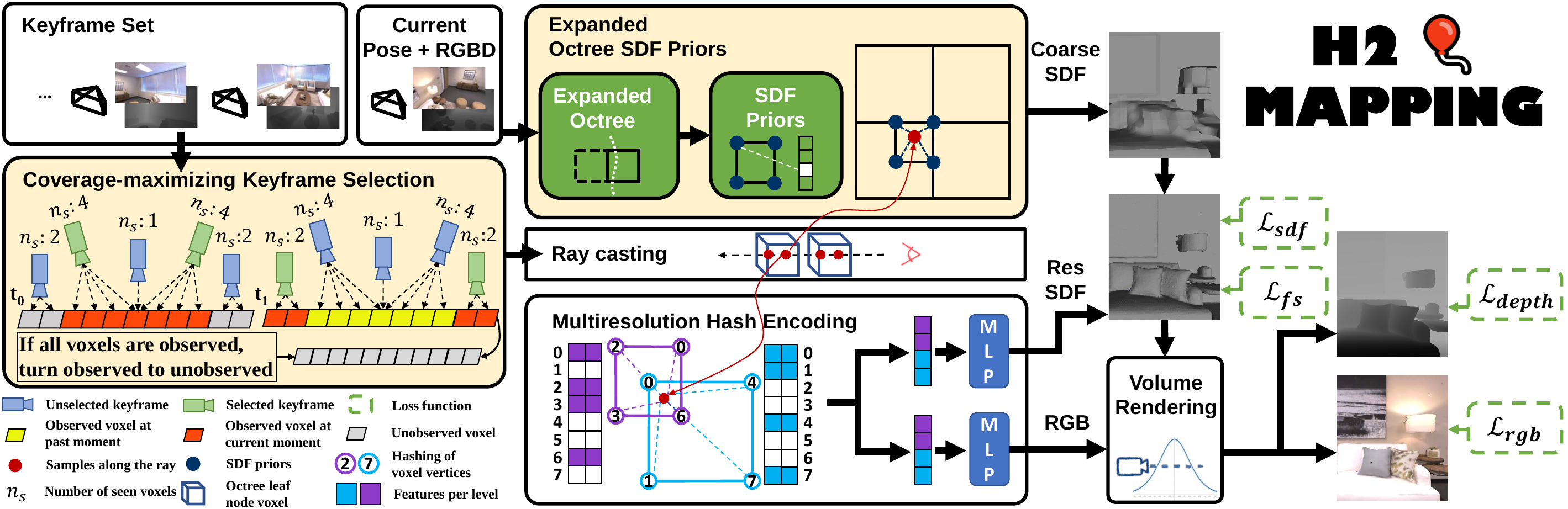}
		\caption{The pipeline of H$_2$-Mapping. Taking RGB-D image from sensors and pose from other tracking modules, we utilize an expanded octree SDF priors and multiresolution hash encoding to represent the scene from rough to detailed. Additionally, the proposed coverage-maximizing keyframe selection strategy ensures quality in the edge regions.}
		\label{fig:pipeline}
		\vspace{-0.6cm}
	\end{figure*}
	
	In robotics, mapping has been studied for decades. Previous works utilize explicit scene representations like occupancy grids\cite{hornung2013octomap}, TSDF \cite{oleynikova2017voxblox,newcombe2011kinectfusion, whelan2012kintinuous, dai2017bundlefusion, niessner2013real, kahler2015very}, surfels\cite{wang2019real, whelan2016elasticfusion}, and meshes\cite{ruetz2019ovpc} to achieve real-time performance. However, these methods face challenges in balancing memory consumption and mapping accuracy\cite{zhong2022shine} and are weak in novel view synthesis. In recent years, implicit representations have gained popularity following the introduction of NeRF\cite{mildenhall2021nerf}. Several works\cite{sucar2021imap, zhu2022nice, yang2022vox} employ NeRF to overcome limitations associated with explicit representations and achieve better mapping results in various aspects. These NeRF-based methods can produce high-fidelity reconstructions using less memory and generate high-quality images from novel views by continuously querying the scene attributes. However, the implicit representation describes the scene as high-dimensional features and neural networks that lack physical meaning, resulting in a long time for training. As a result, these methods cannot run in real-time even on the most powerful edge computers like AGX Orin (as evaluated in Sec.\ref{runtime}).
	
	Aiming to design a real-time and high-quality robot mapping method that fulfills the four requirements mentioned above, we propose a NeRF-based mapping method using a hierarchical hybrid representation. Our approach accelerates the optimization of implicit representation with the aid of an easy-to-optimize explicit representation, describing the scene at different levels of detail. For the coarse scene geometry, we describe it with explicit octree SDF priors. Specifically, we incrementally build a sparse voxel octree with a large voxel size, where we store the optimizable SDF of each leaf node's vertex. To represent geometry details and texture, we use implicit multiresolution hash encoding\cite{muller2022instant} to encode high-resolution scene properties in a memory-efficient way. By using octree SDF priors to capture coarse geometry efficiently, the multiresolution hash encoding can focus solely on the residual geometry, which is much simpler to learn than the complete geometry, thereby improving the geometry accuracy and convergence rate. 
	
	To further speed up, we leverage a simple yet effective method to initialize the octree SDF priors. We project the voxel vertices to the depth image and calculate the associated SDF values. This initialization is based on the observation that a single measurement is usually sufficient to provide a promising estimation of coarse SDF values. Therefore, such a representation can obtain accurate geometry early on, which accelerates the optimization of texture with higher fidelity.
	Besides, to realize higher mapping accuracy, we propose a coverage-maximizing keyframe selection strategy to address the crucial forgetting issue in the online mapping task. Our method avoids redundant sample calculations across all keyframes\cite{sucar2021imap} and ensures quality in marginal areas, without increasing the number of training samples\cite{yang2022vox}.
	
	Our method achieves faster and higher-quality NeRF-based mapping. To summarize, contributions are as follows:
	\begin{itemize}
		\item A hierarchical hybrid representation with an effective initialization technique enables real-time dense mapping with high-fidelity details and dynamical expansion ability, even on edge computers.
		
		\item An effective coverage-maximizing keyframe selection strategy that mitigates the forgetting issue and improves quality, especially in marginal areas.
		
		\item Extensive experiments show our method achieves superior mapping results with less runtime compared to existing NeRF-based mapping methods. To our knowledge, our method is the first to run a NeRF-based mapping method onboard in real-time, as depicted in Fig.\ref{fig:real_world_exp}.
	\end{itemize}
	\setlength{\parskip}{-0.1cm}
	\section{Related Works}
	\subsection{Explicit Dense Mapping} 
	\setlength{\parskip}{-0.1cm}
	% version 2023.5.29 jcx
	Various explicit representations have been used to store scene information for dense mapping.
	Octomap\cite{hornung2013octomap} uses probabilistic occupancy estimation to represent occupied, free, and unknown space. 
	As a pioneer in using SDF for dense mapping, Kinect-Fusion\cite{newcombe2011kinectfusion} leverages volumetric SDF to enable real-time tracking and mapping. Following works improve the scalability\cite{whelan2012kintinuous, niessner2013real}, the efficiency\cite{oleynikova2017voxblox, kahler2015very}, and the global consistency\cite{dai2017bundlefusion}. 
	Moreover,\cite{wang2019real} stores surfel to represent the environment, and\cite{ruetz2019ovpc} represents the robot's surrounding as a watertight 3D mesh.
	These methods are well known for their fast processing speed, which can be attributed to the physical meaning of explicit representations that make them easy to optimize. However, they require large amounts of memory to handle high-detailed mapping\cite{zhong2022shine} and are incapable of realistically rendering from novel views.
	\setlength{\parskip}{-0.1cm}
	\subsection{Implicit Dense Mapping}
	% version 2023.5.29 jcx
	Implicit representations utilize latent features and neural networks to represent a 3D scene in a high-dimensional space. 
	DeepSDF\cite{park2019deepSDF} and Occupancy Networks\cite{mescheder2019occupancy} have shown the potential of implicit representation to model geometry. 
	Recently, NeRF\cite{mildenhall2021nerf} further shows promising results in realistic novel view synthesis from sparse input views. 
	Numerous studies\cite{sucar2021imap, zhu2022nice, yang2022vox, wang2022go} have been inspired by NeRF\cite{mildenhall2021nerf} and utilize implicit representation for incremental dense mapping. These methods achieve more compact and accurate results than explicit representations. 
	The NeRF-based mapping pipeline consists of two main components: (1) Scene representation; and (2) Keyframe selection strategy.
	\subsubsection{Scene representation} 
	\setlength{\parskip}{-0.0cm}
	iMap\cite{sucar2021imap} demonstrates, for the first time, that an MLP can serve as the only scene representation. 
	To overcome the limited representation capacity of a single MLP, NICE-SLAM\cite{zhu2022nice} introduces multi-resolution dense grids to store encoded features of the scene, and MLPs are used to unfold the hidden information. But the pre-allocated grids make NICE-SLAM less scalable and memory inefficient.
	Vox-Fusion\cite{yang2022vox}, instead, only allocates voxels to the area containing the surface, forcing the network to learn more details in those regions. 
	Nonetheless, due to the difficulty in optimizing implicit representations, it is challenging for these methods to meet real-time requirements for robotics applications.
	In contrast, our method utilizes a hierarchical hybrid representation for acceleration and accuracy improvement. This approach enables the implicit representation only to handle the residual geometry and texture, by taking the benefit of explicit structure. Optimizing the residual geometry is generally easier and faster. 
	
	In order to speed up, some previous works aim to accelerate geometry convergence by incorporating geometry priors. INGeo\cite{li2023ingeo}, for instance, scales up the initial density prediction by a factor to increase density as it approaches the surface. However, it requires manual configuration and does not provide a reasonable way to set the scaling factor. Go-surf\cite{wang2022go} initializes its feature grid and geometry decoder to ensure that the initial SDF can represent a sphere centered at the scene origin, but this initialization process cannot adapt to map expansion and has little effect on the observed region. However, due to the hybrid representation, our method can directly initialize the explicit representation by projecting to the input depth image, which can speed up the texture optimization process with higher fidelity by providing accurate geometry in the early stage. Therefore, our method can be deployed to robots for accurate mapping in real time. 
	
	\begin{figure}
		\centering
		\includegraphics[width=1\linewidth]{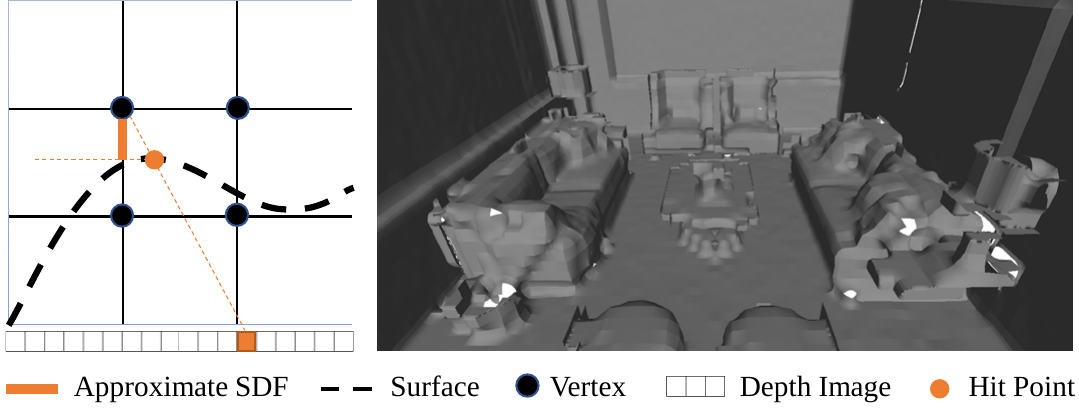}
		\caption{Process of octree SDF priors and the reconstruction results using only the SDF priors without any optimization.}
		\label{fig:sdf_init}
		\vspace{-1.0cm}
	\end{figure}
	\subsubsection{Keyframe selection strategy}  
	iMap\cite{sucar2021imap} allocates samples to every keyframe and calculates the loss distribution for selecting keyframes, which can be redundant. 
	NICE-SLAM\cite{zhu2022nice} selects optimized keyframes based on the overlap with the current frame. This strategy can keep the geometry outside the current field of view static by using a fixed, pre-trained decoder, but it cannot perform well in marginal areas that are seldom observed. 
	Vox-Fusion\cite{yang2022vox} adds a new keyframe based on the ratio of newly allocated voxels to the currently observed voxels. All keyframes are selected to sample the same number of pixels for ray casting, leading to the increasing number of training samples over time.
	However, our coverage-maximizing keyframe selection strategy ensures all allocated voxels are covered with minimal iteration rounds, thereby improving the mapping quality, especially in edge regions.
	\section{H$_2$-Mapping}
	\setlength{\parskip}{-0.1cm}
	In this work, we propose a real-time and high-quality mapping method, as outlined in Fig.\ref{fig:pipeline}. Given a set of sequential poses and RGB-D frames, we utilize a hierarchical hybrid representation (Sec.\ref{subsec:sparse_hierarchical_implicit_representation}) to depict the scene geometry and appearance. By employing a coverage-maximizing keyframe selection strategy (Sec.\ref{subsec:key_slct}), we use the volume rendering approach like NeRF\cite{mildenhall2021nerf} to obtain the depth and color of each sampled ray (Sec.\ref{subsec:sdf-volume-rendering}) and then optimize the hierarchical hybrid representation (Sec.\ref{subsec:optimization_process}).
	\begin{figure}[t]
		\centering
		\includegraphics[width=1\linewidth]{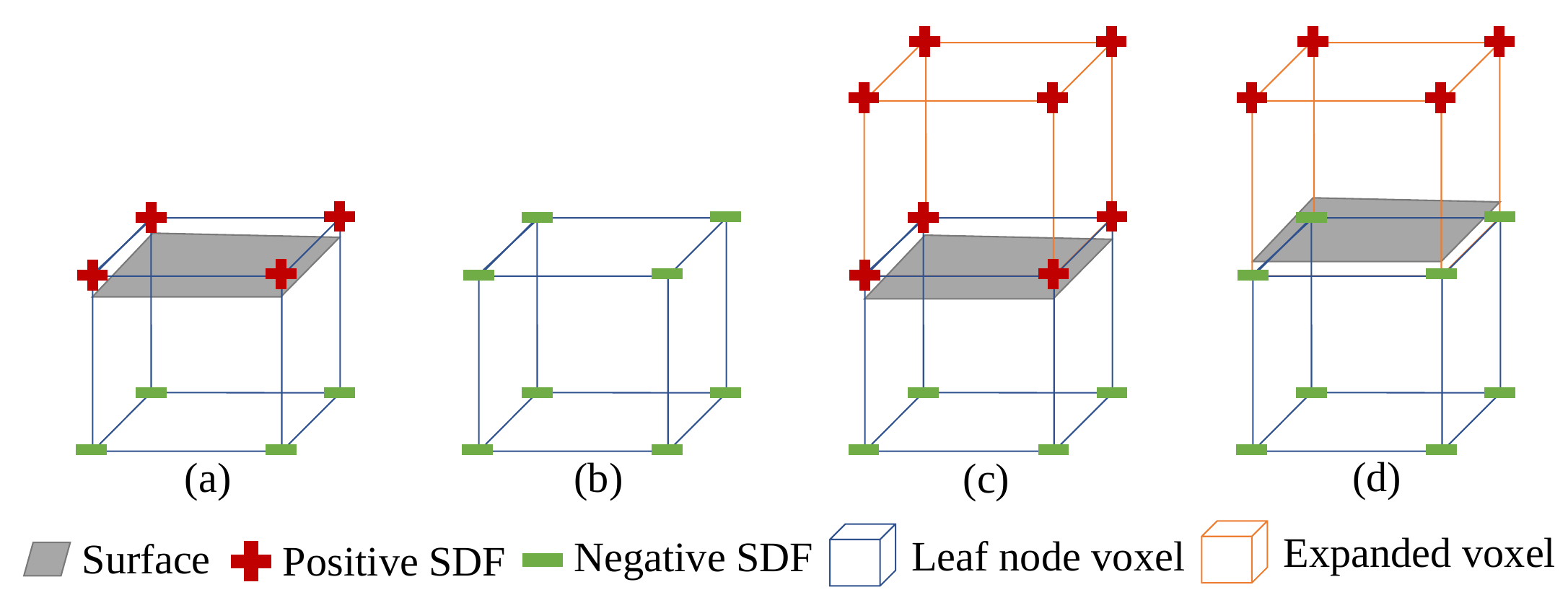}
		\caption{The correct sign of SDF priors (a). And different situations when using voxel expansion (c,d) or not (b) as the surface is near the voxel's boundary.}
		\label{fig:sdf_expand}
		\vspace{-1.2cm}
	\end{figure}
	\setlength{\parskip}{-0.1cm}
	\subsection{Hierarchical Hybrid Representation}
	\setlength{\parskip}{-0.1cm}
	\label{subsec:sparse_hierarchical_implicit_representation}
	To accelerate the optimization of implicit representation, we propose a hierarchical hybrid representation that explicitly stores SDF priors in an expanded octree and uses the implicit multiresolution hash encoding to only handle residual geometry and texture.
	\subsubsection{Expanded Octree SDF Priors}
	\setlength{\parskip}{-0.0cm}
	\label{subsubsec:exp_oct_sdf_priors}
	\paragraph{Octree SDF priors}
	\label{para:oct_sdf_priors}
	When a new frame is received, we allocate new voxels based on the given pose and depth image and incrementally maintain a sparse voxel octree that covers all visible areas. We only add voxels containing more than ten points to the sparse voxel octree to reduce the impact of measurement noise. The number of points is determined through experiments. For each voxel, we store the optimizable SDF in every vertex to represent the coarse geometry of the scene. The coarse SDF $s^c$ of any sample point in a leaf node is obtained from its surrounding eight vertices through the trilinear interpolation function $TriLerp(\cdot)$:
	\begin{equation}
		\begin{aligned}
			s^c = TriLerp(\mathbf{p},  \{s^c_k\}), \quad k \in V,
		\end{aligned}
	\end{equation}
	where $\mathbf{p}$ is the position of the sample point, $s^c_k$ is the optimizable SDF of its surrounding vertex, and $V$ is the set of eight vertices in the leaf node.
	
	To accelerate the convergence rate, we provide an initial SDF to each $s^c_k$ when allocating new voxels. As shown in the left figure of Fig.\ref{fig:sdf_init}, we project every vertex of each voxel onto the corresponding pixel in the RGB-D camera's frame to obtain an approximate SDF at that position:
	\begin{equation}
		\begin{aligned}
			s^c_{prior} = \mathbf{D}(\textbf{u}) - d_{\textbf{p}},
		\end{aligned}
	\end{equation}
	where $d_{\textbf{p}}$ is the z-axis distance between the sensor and the vertex position $\mathbf{p}$, $\mathbf{u}$ is the projected pixel, and $\mathbf{D}(\textbf{u})$ is the depth value at the pixel $\mathbf{u}$. To avoid unreasonable SDF priors due to occlusion, we only provide the prior to the vertices where $ (\mathbf{D}(\mathbf{u}) - d_{\mathbf{p}}) < \sqrt{6}\times \text{(VOXEL SIZE)}$. $\sqrt{6}$ is chosen as the maximum distance between two adjacent voxels. The right figure in Fig.\ref{fig:sdf_init} shows the reconstruction results using only the SDF priors without any optimization. These coarse geometry priors accelerate the geometry optimization and then enhance 
	the scene's appearance by providing accurate geometry in the early stage, which is evaluated in Sec.\ref{eva:sdf_prior}.
	\begin{figure}[t]
		\centering
		\includegraphics[width=1\linewidth]{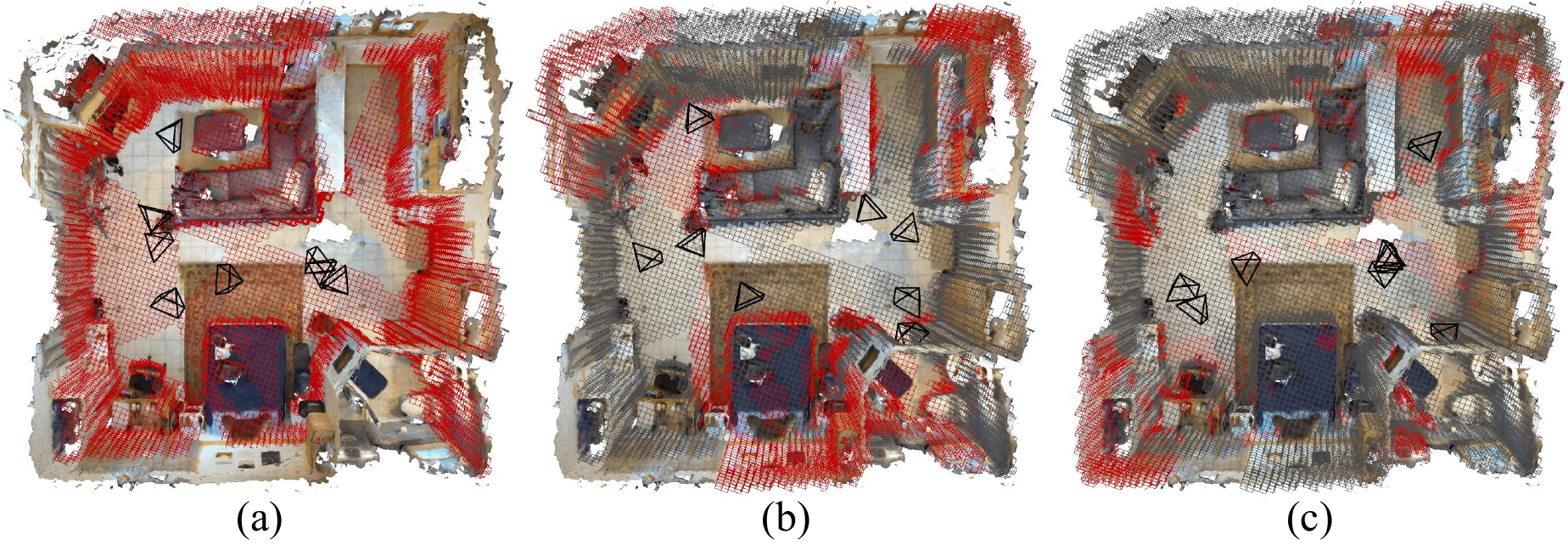}
		\caption{This figure illustrates the voxels covered by the selected keyframes in three consecutive frames. The gray bounding boxes indicate the previously selected voxels, while the red bounding boxes represent the currently chosen voxels. (a) covers most of the voxels in the scene. (b) and (c) covers most of the remaining voxels, such as the toilet and poster areas on the wall.}
		\label{fig:coverage_max_keyframe}
		\vspace{-0.6cm}
	\end{figure}
	\paragraph{Expanded Voxels Allocation}
	If the surface is close to the voxel's boundary, the accurate SDF at the position of the vertex near the surface will be close to 0. Therefore, it is possible for the SDF priors stored in that vertex to be optimized to the wrong sign, leading to the loss of the surface (Fig.\ref{fig:expanding_voxels}(b)). To ensure that a surface will be created, we expand a new voxel if all the points obtained from back-projecting the depth image in the voxel are located at the edge. In Fig.\ref{fig:sdf_expand}, for example, the accurate SDF priors of the upper vertices should be positive but are close to 0 (Fig.\ref{fig:sdf_expand}(a)). Any slight disturbance in the optimization may cause these values to become negative, resulting in no surface being reconstructed (Fig.\ref{fig:sdf_expand}(b)). However, if we allocate an extra voxel on top of it, regardless of the sign to which the vertex near the surface is optimized, a surface will always be built (Fig.\ref{fig:sdf_expand}(c)(d)).
	
	\subsubsection{Multiresolution Hash Encoding}
	\label{subsubsec:multiresolution_hash_encoding}
	In Sec.\ref{subsubsec:exp_oct_sdf_priors}, we efficiently obtain a coarse SDF of the scene. In order to obtain the scene's appearance and more detailed geometry, we employed a multiresolution hash encoding approach inspired by Instant-NGP\cite{muller2022instant}. Differing from the SDF implementation in Instant-NGP\cite{muller2022instant}, we only utilize the multiresolution hash encoding to handle the residual SDF which is easy to learn than the complete SDF of the scene. 
	The multiresolution hash encoding works by arranging the surrounding voxels of a particular sample point at $L$ resolution levels. At each level, $F$ dimensional features are assigned to the corners of the voxels by looking up a hash table. To obtain the feature of the sample point, tri-linear interpolation is performed, and the feature at each level is concatenated. We employ two multiresolution hash encoding and shallow MLP attached to individually represent the color and residual SDF of the scene in a compact manner:
	\begin{equation}
		\begin{aligned}
			s = s^{c} + \mathcal{M}_s(\phi ^{s};\theta^{w}_{s}), \quad
			\mathbf{c} = \mathcal{M}_c(\phi ^{c};\theta^{w}_{c}),
		\end{aligned}
	\end{equation}
	where $\phi ^{c}$ and $\phi ^{s}$ are $L \times F$ dimensional features obtained from the multiresolution hash encoding. $\mathcal{M}_s$ and $\mathcal{M}_c$, parameterized by $\theta^{w}_{s}$ and  $\theta^{w}_{c}$, are MLPs to output the residual SDF prediction $s$ and color prediction $\mathbf{c}$ ($\mathbb{R}^3$ for $RGB$ respectively).

	\subsection{Coverage-maximizing Keyframe Selection}
	\label{subsec:key_slct}
	For a new input RGB-D frame, we insert this frame as a new keyframe if the ratio $(N_{c} \cap N_{l})/(N_{c} \cup N_{l})$ is smaller than a threshold, where $N_{c}$ is the number of currently observed voxels, $N_{l}$ is the number of voxels observed at the last inserted keyframe. Our keyframe insertion strategy ensures that the frames in the keyframe set have relatively little overlap.
	
	% 根据上述的关键帧插入策略，关键帧观测到的体素包含了场景的大部分区域。 所以我们提出将关键帧选择定义为覆盖最大化体素。
	To select the optimized keyframes from the keyframe set, we employ a coverage-maximizing keyframe selection strategy, as illustrated in Fig. \ref{fig:pipeline}. At the initial time step $t_0$, all voxels are labeled as unobserved. We begin by selecting $K$ keyframes that cover the largest number of voxels from the entire keyframe set. We mark these covered voxels as observed, and then optimize these selected keyframes and the current frame jointly. In the next time step $t_1$, we use the same coverage-maximizing strategy but only for voxels that are still labeled as unobserved. If all voxels have been labeled as observed, we reset the voxels that were previously marked as observed to unobserved and repeat the above process. By using this strategy iteratively, all the scene areas can be covered. As shown in Fig. \ref{fig:coverage_max_keyframe}(a), most of the voxels are covered in $t_0$. In Fig. \ref{fig:coverage_max_keyframe}(b) and (c), the strategy continues to cover other remaining parts of the scene, ensuring the reconstruction quality of the edge regions. In Sec.\ref{eva:keyframe}, we further evaluate this strategy.

	\begin{table*}[]
		\centering
		\caption{Reconstruction results of 8 scenes in the Replica dataset\cite{straub2019replica}. Compared with NICE-SLAM\cite{zhu2022nice} and Vox-Fusion\cite{yang2022vox}, our approach yields better results in all the metrics.}
		\label{tab::replica_recon_render_metric}
		\begin{tabular}{c||c||ccccccccc}
			\Xhline{2\arrayrulewidth}
			Metrics                                                                       & Method              & Room0          & Room1          & Room2          & Office0        & Office1        & Office2        & Office3        & Office4        & Avg.           \\ \hline
			\multirow{3}{*}{Depth L1${[}cm{]}$ $\downarrow$ }                                             & NICE-SLAM           & 1.19           & 0.93           & 1.28           & 1.02           & 1.35           & 1.07           & 1.30           & 1.23           & 1.16           \\
			& Vox-Fusion          & 0.51           & 0.47           & 0.95           & 0.65           & 0.87           & 0.71           & 0.81           & 0.72           & 0.71           \\
			& \textbf{H$_2$-Mapping} & \textbf{0.34}  & \textbf{0.22}  & \textbf{0.61}  & \textbf{0.33}  & \textbf{0.45}  & \textbf{0.53}  & \textbf{0.50}  & \textbf{0.40}  & \textbf{0.42}  \\ \hline
			\multirow{3}{*}{Acc.${[}cm{]}$ $\downarrow$ }                                                 & NICE-SLAM           & 1.56           & 1.28           & 1.40           & 1.20           & 1.03           & 1.44           & 1.93           & 1.61           & 1.43           \\
			& Vox-Fusion          & 1.28           & 1.09           & 1.17           & 1.04           & 1.13           & 1.27           & 1.55           & 1.43           & 1.25           \\
			& \textbf{H$_2$-Mapping} & \textbf{1.25}  & \textbf{0.99}  & \textbf{1.09}  & \textbf{0.97}  & \textbf{0.84}  & \textbf{1.17}  & \textbf{1.38}  & \textbf{1.29}  & \textbf{1.12}  \\ \hline
			\multirow{3}{*}{Comp.${[}cm{]}$ $\downarrow$ }                                                & NICE-SLAM           & 1.62           & 1.34           & 1.63           & 1.12           & 1.24           & 1.59           & 2.31           & 1.78           & 1.58           \\
			& Vox-Fusion          & 1.32           & 1.10           & 1.25           & 1.09           & 1.19           & 1.26           & 1.54           & 1.45           & 1.27           \\
			& \textbf{H$_2$-Mapping} & \textbf{1.26}  & \textbf{0.97}  & \textbf{1.13}  & \textbf{0.98}  & \textbf{0.84}  & \textbf{1.16}  & \textbf{1.38}  & \textbf{1.34}  & \textbf{1.13}  \\ \hline
			\multirow{3}{*}{Comp. Ratio${[}\textless{}5cm\%{]}$ $\uparrow$}                            & NICE-SLAM           & 98.41          & 98.74          & 97.22          & 95.15          & 97.37          & 97.61          & 94.89          & 97.35          & 97.09          \\
			& Vox-Fusion          & 98.86          & 99.27          & 98.28          & 98.74          & 98.18          & 98.90          & 97.84          & 98.35          & 98.55          \\
			& \textbf{H$_2$-Mapping} & \textbf{99.28} & \textbf{99.80} & \textbf{99.14} & \textbf{99.17} & \textbf{99.25} & \textbf{99.53} & \textbf{98.88} & \textbf{98.98} & \textbf{99.25} \\ \hline
			\multirow{3}{*}{\begin{tabular}[c]{@{}c@{}}SSIM (Interpolate) $\uparrow$ \end{tabular}} & NICE-SLAM           & 0.92           & 0.93           & 0.93           & 0.96           & 0.97           & 0.94           & 0.94           & 0.95           & 0.94           \\
			& Vox-Fusion          & 0.93           & 0.94           & 0.95           & 0.97           & 0.96           & 0.95           & 0.95           & 0.96           & 0.95           \\
			& \textbf{H$_2$-Mapping} & \textbf{0.95}  & \textbf{0.97}  & \textbf{0.97}  & \textbf{0.99}  & \textbf{0.98}  & \textbf{0.97}  & \textbf{0.97}  & \textbf{0.98}  & \textbf{0.97}  \\ \hline
			\multirow{3}{*}{\begin{tabular}[c]{@{}c@{}}PSNR$[db]$ (Interpolate) $\uparrow$\end{tabular}} & NICE-SLAM           & 27.19          & 28.88          & 30.28          & 34.14          & 34.82          & 29.11          & 29.43          & 30.51          & 30.55          \\
			& Vox-Fusion          & 29.38          & 30.95          & 30.29          & 33.64          & 33.63          & 30.74          & 30.63          & 32.58          & 31.48          \\
			& \textbf{H$_2$-Mapping} & \textbf{31.76} & \textbf{33.61} & \textbf{32.89} & \textbf{38.67} & \textbf{38.92} & \textbf{32.68} & \textbf{33.13} & \textbf{34.34} & \textbf{34.49} \\ \hline
			\multirow{3}{*}{\begin{tabular}[c]{@{}c@{}}SSIM (Extrapolate) $\uparrow$\end{tabular}} & NICE-SLAM           & 0.92           & 0.92           & 0.92           & 0.96           & 0.92           & 0.93           & 0.94           & 0.95           & 0.93           \\
			& Vox-Fusion          & 0.92           & 0.92           & 0.93           & 0.96           & 0.94           & 0.94           & 0.93           & 0.95           & 0.94           \\
			& \textbf{H$_2$-Mapping} & \textbf{0.95}  & \textbf{0.95}  & \textbf{0.95}  & \textbf{0.97}  & \textbf{0.96}  & \textbf{0.95}  & \textbf{0.96}  & \textbf{0.96}  & \textbf{0.95}  \\ \hline
			\multirow{3}{*}{\begin{tabular}[c]{@{}c@{}}PSNR$[db]$ (Extrapolate) $\uparrow$\end{tabular}} & NICE-SLAM           & 26.96          & 27.99          & 26.67          & 32.21          & 30.10          & 25.59          & 26.66          & 28.06          & 28.08          \\
			& Vox-Fusion          & 25.90          & 26.31          & 26.03          & 30.86          & 29.08          & 25.91          & 26.49          & 28.17          & 27.34          \\
			& \textbf{H$_2$-Mapping} & \textbf{29.24} & \textbf{28.35} & \textbf{27.05} & \textbf{33.72} & \textbf{33.82} & \textbf{28.91} & \textbf{29.43} & \textbf{31.17} & \textbf{30.21} \\ \Xhline{2\arrayrulewidth}
		\end{tabular}
	\vspace{-1.5cm}
	\end{table*}
	
	\subsection{SDF-based Volume rendering}
	\label{subsec:sdf-volume-rendering}
	Like Vox-Fusion\cite{yang2022vox}, we only sample points along the ray that intersects with any voxel. And then get rendered color $\mathbf{C}$ and depth $D$ for each ray as follows:
	\begin{equation}
		\begin{aligned}
			w_{j} &= \sigma(\frac{s_j}{tr}) \cdot \sigma(-\frac{s_j}{tr}) \\
			\mathbf{C} &= \frac{1}{\sum^{N-1}_{j=0}w_j}\sum^{N-1}_{j=0}w_j\cdot\mathbf{c}_j, \  
			D = \frac{1}{\sum^{N-1}_{j=0}w_j}\sum^{N-1}_{j=0}w_j\cdot d_j, 
		\end{aligned}
	\end{equation}
	where $\sigma(\cdot)$ is the sigmoid function, $s_{j}$ and $\mathbf{c}_j$ are the predicted SDF and color obtained from the hierarchical hybrid representation described in Sec. \ref{subsec:sparse_hierarchical_implicit_representation}, $N$ is the number of samples along the ray, $tr$ is a truncation distance and $d_{j}$ is the sample's depth along the ray.
	
	\subsection{Optimization Process}
	\label{subsec:optimization_process}
	\subsubsection{Loss Function}
	We apply loss functions like Vox-Fusion\cite{yang2022vox}: RGB Loss ($\mathcal{L}_{rgb}$), Depth Loss ($\mathcal{L}_{d}$), Free Space Loss ($\mathcal{L}_{fs}$) and SDF Loss ($\mathcal{L}_{sdf}$) on a batch of rays $R$. 
	\begin{equation}
		\resizebox{0.98\hsize}{!}{$
			\begin{aligned}
				\mathcal{L}_{\!fs\!}\! &=\! \frac{1}{\vert R\vert }\!\sum_{\!r\in R}\! \frac{1}{P_r^{fs}}\!\sum_{p \in P_r^{fs}}\!(s_p\! -\! tr\!)^2 , \ 
				\mathcal{L}_{\!sdf\!}\! =\! \frac{1}{\vert R\vert }\!\sum_{\!r\in R}\! \frac{1}{P_r^{tr}}\!\sum_{p \in P_r^{tr}}\!(s_p\! -\! s_p^{gt}\!)^2 \\
				\mathcal{L}_{\!d\!} &= \frac{1}{\vert R\vert}\!\sum_{\!r\in R}\lVert D_r\! -\! D^{gt}_r\rVert , \ 
				\mathcal{L}_{\!rgb\!} = \frac{1}{\vert R\vert}\!\sum_{\!r\in R}\lVert \mathbf{C}_r\! -\! \mathbf{C}^{gt}_r\rVert
			\end{aligned}$}
	\end{equation}
	where $P_r^{fs}$ is a set of points on the ray $r$ that lies between the camera and the truncation region of the surface measured by the depth sensor, $P_r^{tr}$ is a set of points within the truncation area. ($D_{r}$, $D^{gt}_r$) and ($\textbf{C}_{r}$, $\mathbf{C}^{gt}_r$) are rendered and input depth and color. $s_p$ is the predicted SDF and $s_p^{gt}$ is the difference between the distance to point $p$ on the ray $r$ and the depth measurement of that ray.
	The final loss function is defined as:
	\begin{equation}
		\begin{aligned}
			\mathcal{L} = \alpha_{sdf}\mathcal{L}_{sdf} + \alpha_{fs}\mathcal{L}_{fs} +\alpha_{d}\mathcal{L}_{d} +\alpha_{rgb}\mathcal{L}_{rgb}, 
		\end{aligned}
	\end{equation}
	where $\alpha_{sdf}$, $\alpha_{fs}$ ,$\alpha_{d}$, and $\alpha_{rgb}$ are the weighting coefficients.
	\setlength{\parskip}{-0.1cm}
	\subsubsection{Adaptive Early Ending}
	In the training process, the current frame and selected keyframes will be used for optimization several times. As shown in Fig. \ref{fig:iters_ending}, the average iteration time that achieves color optimization convergence varies in different scenarios. Therefore, to adaptively choose an appropriate iteration time that can balance the time consumption and mapping precision in various scenarios, we employ an early stopping policy if the total loss exceeds twice the average total loss of the current training round, indicating that further optimization can only result in a little improvement.
	\section{Experiment}
	To evaluate the performance of our proposed method, we compare its mapping accuracy and time consumption with other NeRF-based RGB-D mapping systems on both the synthetic Replica dataset\cite{straub2019replica} and the real-world ScanNet dataset\cite{dai2017scannet}. Besides, we conduct ablation studies to evaluate the effectiveness of each module in our approach. Furthermore, we deploy our method on a handheld device and quadrotors with limited computational power to test its mapping performance.
	\subsection{Mapping and Rendering Evaluation}
	% \subsection{Experimental Setup}
	% \subsubsection{Datasets}
	% Our experiments independently select synthetic and real sequence datasets. For the simulated dataset, we utilize the Replica dataset\cite{straub2019replica} for quantitative analysis. Due to the inaccuracy of the ground truth of the real scene dataset, we chose Scannet dataset\cite{dai2017scannet} as the real sequence dataset but only for qualitative analysis.
	\begin{figure*}[h!]
		\centering
		\includegraphics[width=1.0\textwidth]{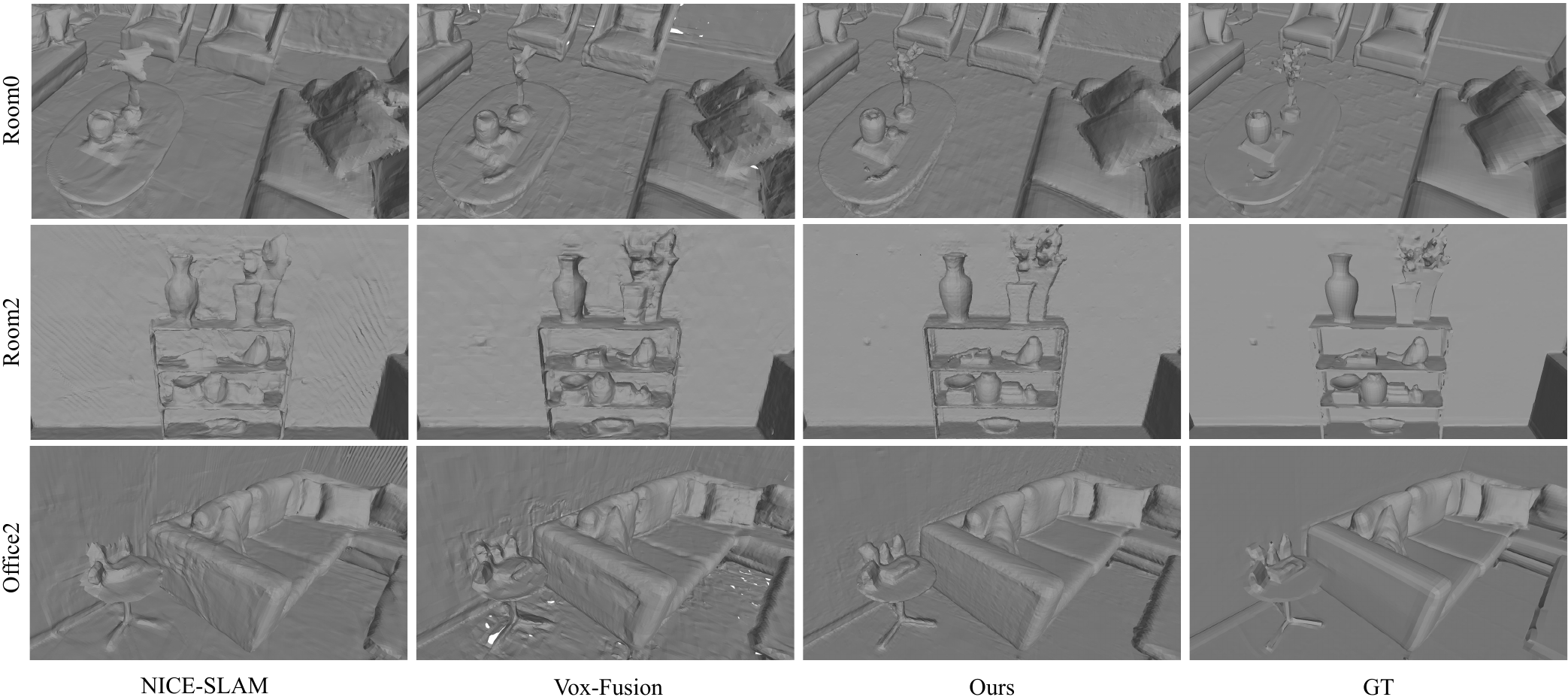}
		\caption{Reconstructed mesh on the Replica dataset\cite{straub2019replica}. Our method closely approximates the ground truth mesh and can capture richer geometric details in small objects, such as chair legs and vases.}
		\label{fig:replica_mesh_image}
	\end{figure*}
	\begin{figure*}[h!]
		\centering
		\includegraphics[width=1.0\textwidth]{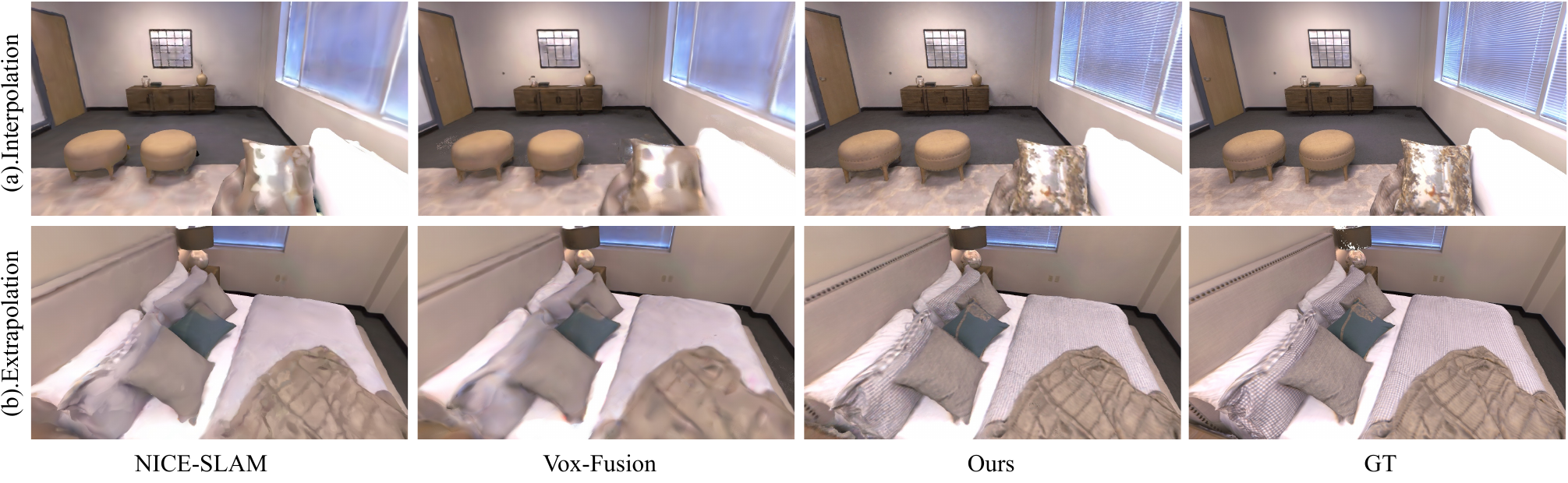}
		\caption{Image rendering results on the Replica dataset\cite{straub2019replica}. (a) Rendering results from the input views (interpolation). (b) Rendering results from the novel views (extrapolation). The image rendered by our method has more high-fidelity details and rich textures, such as pillows and bed sheets}
		\label{fig:replica_render_image}
	\end{figure*}
	\begin{figure*}[h!]
		\centering
		\includegraphics[width=1.0\textwidth]{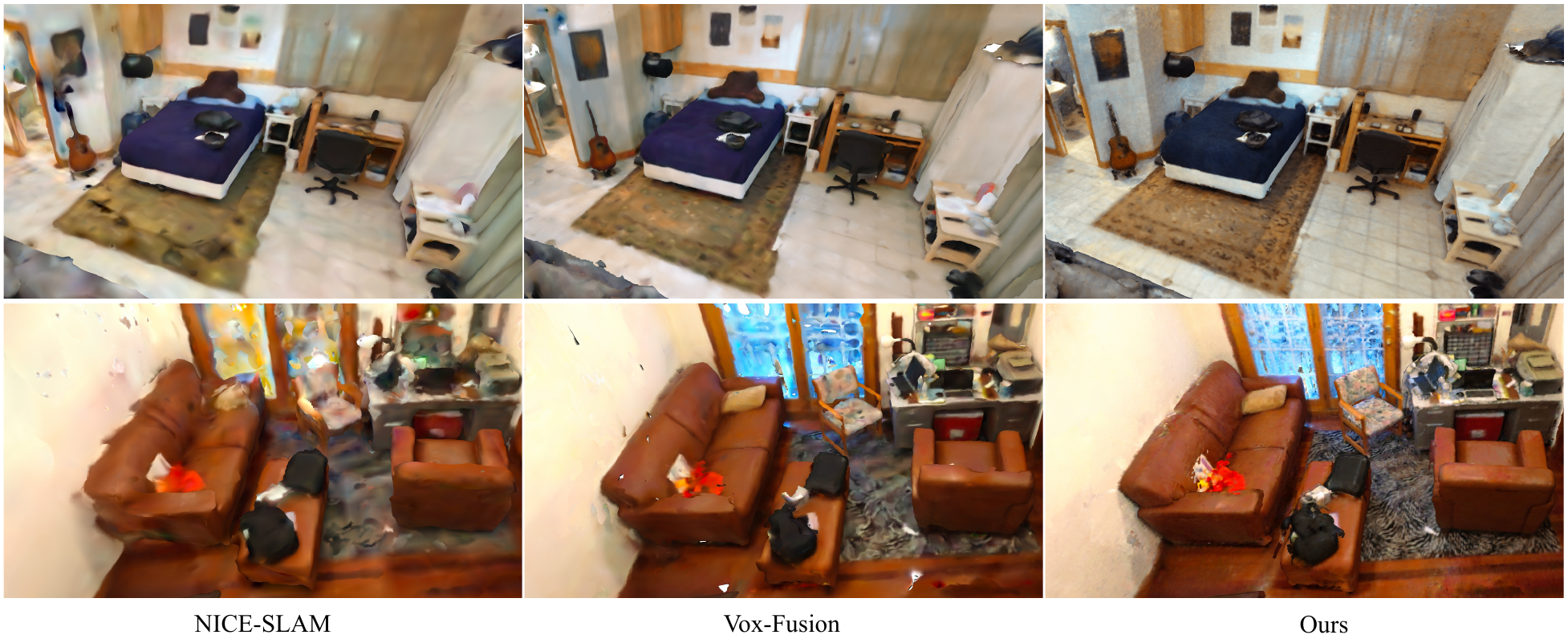}
		\caption{Textured mesh on the ScanNet dataset\cite{dai2017scannet}. Our method produces more accurate geometry and higher-fidelity textures.}
		\label{fig:scannet_results}
		\vspace{-0.8cm}
	\end{figure*}
	\subsubsection{Implementation Details} In our method, the voxel size of the octree's leaf node is $10cm$, $tr=5cm$, and the maximum number of iterations is $10$. For the multi-resolution hash encoding, $L=4$, $F=2$, $T=2^{19}$, and the scale difference of the adjacent level is $2$. Due to octree SDF priors, we only use one-layer MLP of size $64$ to decode the geometry features. The appearance decoder is a two-layer MLP of size $64$. $4096$ pixels are selected for each iteration to generate rays and the distance between adjacent sampled points is $1cm$. The number of keyframes that are selected to be optimized is $K = 10$.
	% We use the same set of loss coefficients for mapping. We set $\lambda_{SDF}=50000$, $\lambda_{fs}=10$, $\lambda_{d}=1$, $\lambda_{c}=0.5$.
	
	\subsubsection{Baselines} 
	We select two advanced NeRF-based dense RGB-D SLAM methods currently open-source, NICE-SLAM\cite{zhu2022nice} and Vox-Fusion\cite{yang2022vox} for comparison. However, since we solely focus  on incremental mapping, we remove their tracking component and instead provide the ground truth pose. All other aspects remain unchanged.
	
	\subsubsection{Metrics} To evaluate scene geometry, we use the Depth L1 Error$[cm]$ Accuracy$[cm]$, Completion$[cm]$ and Completion Ratio$[<5cm\%]$ of the reconstructed mesh. 
	Besides, We use SSIM (structural similarity index measure)\cite{wang2004image} and PSNR (Peak signal-to-noise ratio)\cite{hore2010image} like to evaluate scene appearance on rendered images from all training views (Interpolation) and distant novel views (Extrapolation). 
	Only the portions of the mesh included in voxels are considered, and non-depth regions are not measured in Depth L1 Error, SSIM\cite{wang2004image}, and PSNR\cite{hore2010image}.
	
	% For geometry evaluation, we compare the reconstructed mesh where voxels have occupied with the GT mesh like NICE-SLAM\cite{zhu2022nice}, We consider \textit{Accuracy}[cm], \textit{Completion}[cm], \textit{Completion Ratio}[$<$5cm\%] for the 3D metrics and \textit{Depth L1 Error[mm]} for the 2D metrics. We perform mesh culling by retaining only the voxels at the leaf node positions of the octree. For evaluating scene appearance, we borrow from NICER-SLAM\cite{zhu2023nicer}. We assess the extent to which the sequence captures appearances by rendering images from all training viewpoints(\textbf{Interpolate}). To evaluate the effectiveness of a scene in rendering the appearance of a novel viewpoint, we select distant novel viewpoints from the input view(\textbf{Extrapolation}). Moreover, we use both SSIM\cite{wang2004image} and PSNR metrics.

	% For the 2D metric, we render depth maps from 1000 random camera poses in each scene and calculate the L1[mm] difference between depths from ground truth meshes and the reconstructed ones. 
	
	\subsubsection{Evaluation on Replica\cite{straub2019replica}}
	In \autoref{tab::replica_recon_render_metric}, we present a quantitative comparison of the reconstruction and rendering performance of our method and the baselines. The results demonstrate that our approach outperforms the baselines for both 2D and 3D metrics. Additionally, we provide a qualitative analysis of the reconstructed mesh and rendered images. Notably, in Fig. \ref{fig:replica_mesh_image}, the mesh obtained by our method appears smoother in areas such as the sofa and floor, while our approach exhibits enhanced geometric details, particularly for smaller objects like chair legs and vases. Fig. \ref{fig:replica_render_image} shows that our method can generate renderings with more realistic details from both training views and novel views. 
	
	\subsubsection{Evaluation on ScanNet\cite{dai2017scannet}}
	Since ground truth meshes for the ScanNet\cite{dai2017scannet} are unavailable, we only provide qualitative analysis on the textured mesh of Scene0000 and Scene0050 as shown in Fig. \ref{fig:scannet_results}. Compared to the baselines, our methods can get sharper object outline, fewer artifacts, and higher-fidelity textures like the pattern on the carpet.
	
	\begin{figure}[t!]
		\centering
		\includegraphics[width=\linewidth]{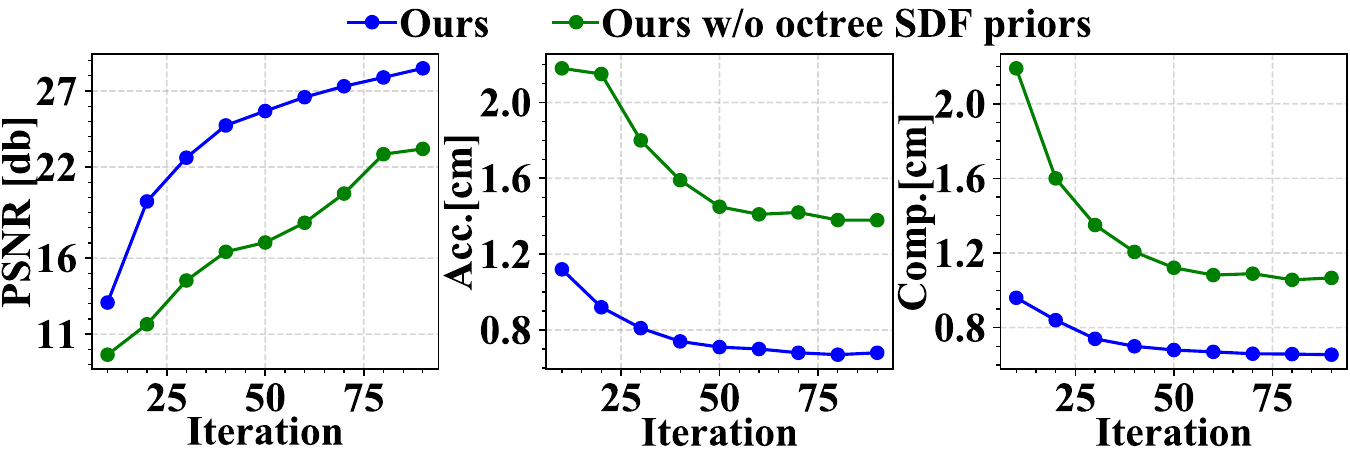}
		\caption{Ablation study about octree SDF priors on one frame. We choose the first frame in Room0 of Replica\cite{straub2019replica} to evaluate how the mapping performance of the corresponding region changes with increasing optimization iterations.}
		\label{fig:sdf_init_one_frame}
		\vspace{-0.2cm}
	\end{figure}
	\subsubsection{Runtime Analysis}
	\label{runtime}
	We select Room0 from Replica\cite{straub2019replica} and Scene0000 from ScanNet\cite{dai2017scannet} to evaluate the runtime by comparing our method with the baselines in \autoref{tab:timetable}. We report the average frame processing time (FPT) on RTX 4090, RTX 2080Ti, AGX Orin, Orin NX separately. Our method is much faster than previous work on various devices.
	
	% Our method demonstrates real-time performance on an X86 graphics computing platform and near-real-time performance on an ARM device (Jetson AGX Orin and Jetson Orin NX). 
	
	\begin{table}[]
		\centering
		\caption{Runtime Analysis on Multiple Devices}
		\label{tab:timetable}
		\resizebox{\linewidth}{!}{
			\begin{tabular}{c||c||cccc}
				\Xhline{2\arrayrulewidth}
				\multicolumn{1}{l||}{\multirow{2}{*}{}} & \multirow{2}{*}{Method} & \multicolumn{4}{c}{Speed FTP(s) $\downarrow$}                               \\ \cline{3-6} 
				\multicolumn{1}{l||}{}                  &                         & \multicolumn{1}{l}{4090} & \multicolumn{1}{l}{2080Ti} & AGX Orin & Orin NX \\ \hline
				\multirow{3}{*}{\makecell{Replica\\Room0}}               
				& NICE-SLAM               & 2.69                       &   6.35                        &13.01          &19.03   \\
				& Vox-Fusion              & 0.34                       &       0.94                     & 1.68          & 3.52   \\
				& \textbf{H$_2$-Mapping}           & \textbf{0.05}              & \textbf{0.16}                  & \textbf{0.36} & \textbf{0.66}\\ \hline
				\multirow{3}{*}{\makecell{ScanNet\\Scene0000}}               
				& Nice-slam               & 3.03                      &  6.55                          & 14.37         & 19.39   \\
				& Vox-Fusion              & 0.67                       &  2.04                          & 3.12   & 21.57\\
				& \textbf{H$_2$-Mapping}           & \textbf{0.07}              &  \textbf{0.24}                 & \textbf{0.53} & \textbf{1.07} \\ 
				\Xhline{2\arrayrulewidth}
			\end{tabular}
		}
		\vspace{-0.5cm}
	\end{table}

	% \usepackage{multirow}
	% \usepackage{multirow}
	% \usepackage{multirow}
	% \usepackage{multirow}
	% \usepackage{multirow}

	\subsection{Ablation Study}

	%\subsubsection{Ablation Study Of Our Design} 
	% We conduct multiple experiments in \autoref{tab:ablation} to defend our design choices in our approach. The comprehensive experiment is conducted on Replica's room0 scene to rigorously validate the efficacy of each individual module. The details are as follows:(a)We only use multi-resolution hash coding as a benchmark, then join each module individually to test validity.(b) We add the SDF initialization method to provide prior information for the appearance representation. (c) We opt to optimize the values of the eight vertices in the octant leaf nodes of the voxel representation. (d) We employ the expanded voxels allocation module in our method. (e) We implement the keyframe strategy. (f) We evaluate our full model. Note that the input to our appearance indicator selection is the interpolated view angle.
	
	\subsubsection{Octree SDF Priors}
	\label{eva:sdf_prior}
	Octree SDF priors represent an easy-to-optimize explicit structure with a computationally efficient initialization process. As a result, the reconstructed geometry depicted in Fig. \ref{fig:sdf_init_one_frame} is more accurate at the beginning of optimization, leading to a rapid PSNR\cite{hore2010image} increase for the first frame during the early stage. Furthermore, by promptly providing a coarse geometry and utilizing the implicit multiresolution hash encoding exclusively for the residual part, the accuracy and completion metrics in Fig. \ref{fig:sdf_init_one_frame} eventually converge to a lower level, leading to higher PSNR\cite{hore2010image} for the same average iteration times on the entire sequence in Fig. \ref{fig:iters_ending}. The PSNR\cite{hore2010image} arises because accurate geometry ensures the gradient of color prediction mainly affects the surface region during backpropagation. Therefore, by enabling a faster and more accurate geometry reconstruction, this hybrid representation achieves the same reconstruction quality with fewer training iterations. This is particularly meaningful for robotic applications with limited computing power.
	
	% As shown in Fig. \ref{fig:iters_ending}, we compared our method without octree SDF priors. We found that the PSNR is lower for the same number of iterations. It shows that octree SDF priors can help the network converge faster and accelerate the training process. According to the principles of volume rendering, geometry influences the color rendering. We demonstrated that our octree SDF prior can accelerate color convergence, which is challenging to optimize. Due to the computational efficiency of SDF priors, we can achieve relatively good mesh quality with lower iteration numbers on devices with limited computing resources. This is particularly meaningful for robotic applications.
	\setlength{\parskip}{-0.0cm}
	\subsubsection{Expanded Voxels Allocation} 
\begin{figure}[t!]
		\centering
		\includegraphics[width=\linewidth]{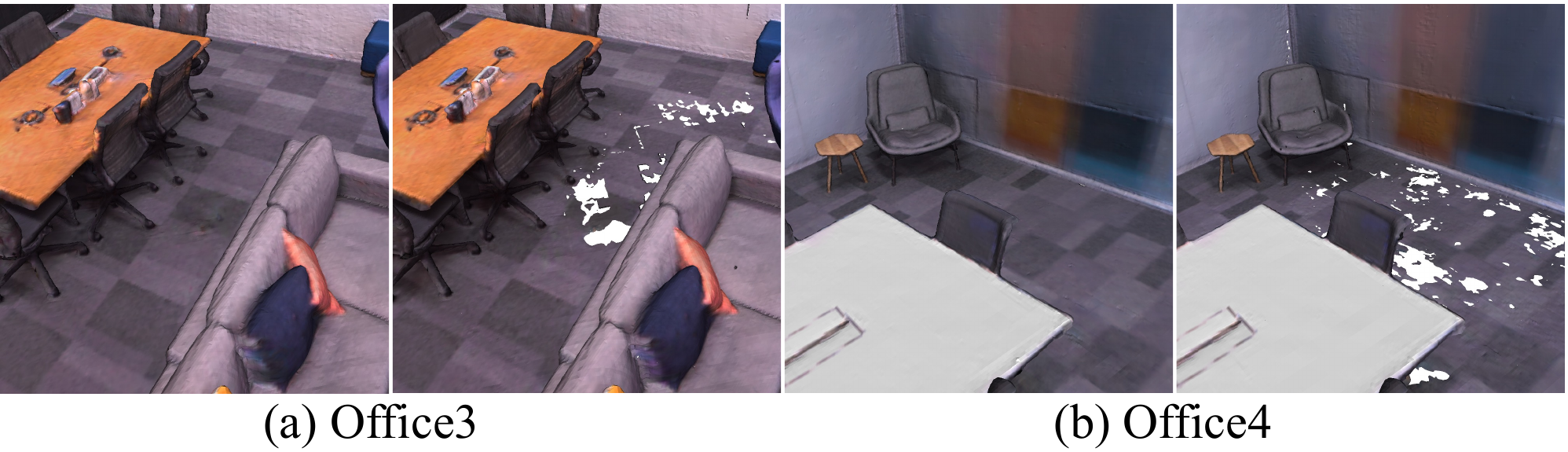}
		\caption{Ablation study on expanded voxels allocation technique. (a) and (b) show results about Office3 and Office4 of Replica\cite{straub2019replica}, respectively. This technique can produce fewer holes during the mapping (Use: left; No use: right).}
		\label{fig:expanding_voxels}
		\vspace{-0.4cm} 
	\end{figure}

Table \ref{tab:ablation} shows the expanded voxel allocation technique has a greater impact on completion. Besides, Fig. \ref{fig:expanding_voxels} shows the visualization results about Office3 and Office4 of Replica\cite{straub2019replica}. In the left part of (a) and (b), holes are generated in regions where the surface is close to the voxel's boundary, making it easy for the SDF to be optimized to the wrong sign. However, as shown in the right part of (a) and (b), the expansion technique can reduce the holes caused by optimization sensitivity.
	
	\subsubsection{Coverage-maximizing Keyframe Selection} 
	\label{eva:keyframe}
\begin{figure}[t!]
		\centering
		\includegraphics[width=\linewidth]{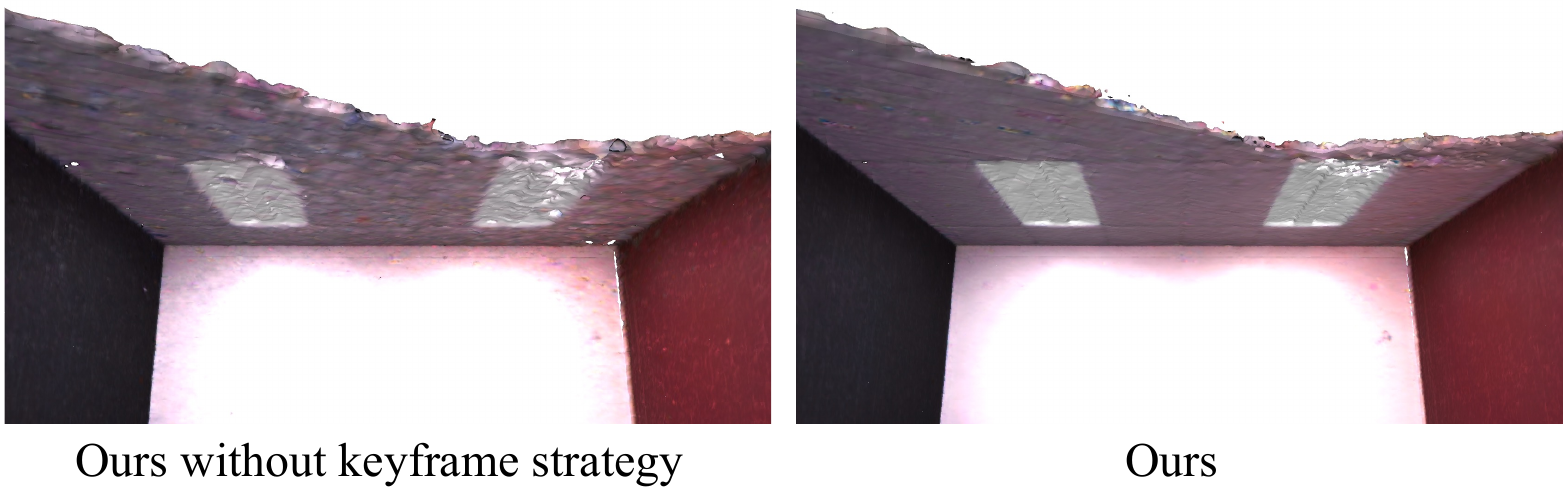}
		\caption{Ablation study of coverage-maximizing keyframe selection strategy on Office2 of Replica\cite{straub2019replica}. It shows that our keyframe selection strategy will lead to better reconstruction performance in marginal areas such as the ceiling.}
		\label{fig::keyframe}
		\vspace{-2.2cm} 
	\end{figure}

	~\autoref{tab:ablation} demonstrate our keyframe selection strategy can significantly improve the PSNR\cite{hore2010image} and Acc. metrics. As shown in Fig. \ref{fig::keyframe}, our strategy can better optimize the ceiling area in Office2 of Replica\cite{straub2019replica}. Since the number of images corresponding to this area is low in the overall keyframe set, it is rare for previous methods to optimize this region using the random strategy. However, our keyframe selection strategy achieves complete coverage of all the voxels in the keyframe set with minimal iteration rounds, greatly increasing the probability of optimizing the edge regions.
	\begin{figure}[t!]
		\includegraphics[width=\linewidth]{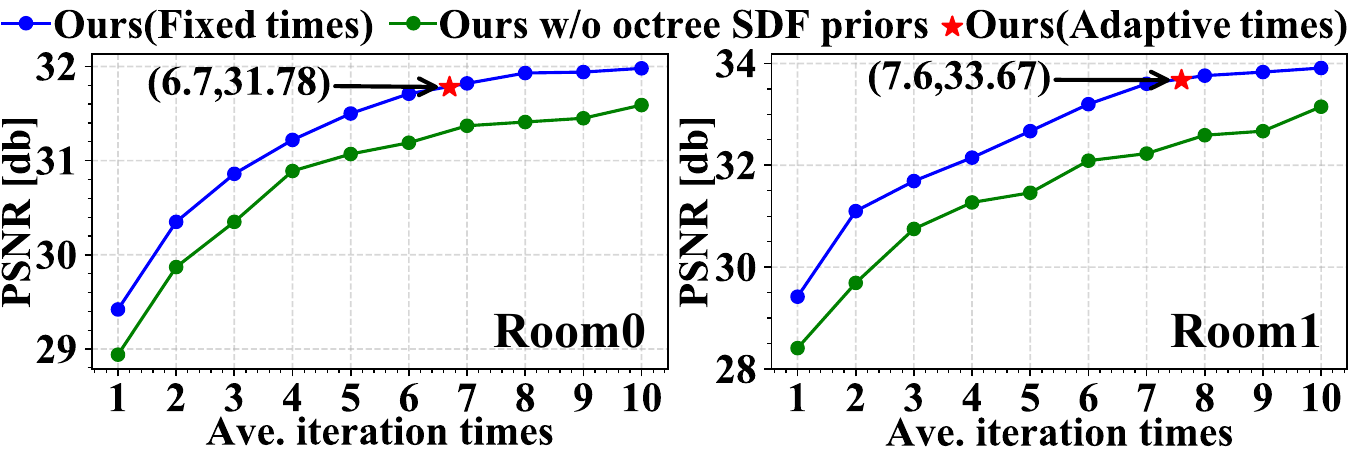}
		\caption{Ablation study about the adaptive early ending and octree SDF priors on the entire data sequence. We compared changes in the PSNR\cite{hore2010image} for different fixed iteration times in Room0 (left) and Room1 (right) of Replica\cite{straub2019replica}. The \textcolor{blue}{blue curve} and \textcolor{Green}{green curve} represent the results with and without octree SDF priors, respectively. The \textcolor{Red}{red star} denotes the average iteration time and PSNR\cite{wang2004image} using adaptive early ending.
		}
		\label{fig:iters_ending}
		\vspace{-0.3cm}   
	\end{figure}
	% \begin{figure}[t!]
		% 	\includegraphics[width=\linewidth]{Image/room0_room1_new3.pdf}
		% 	\caption{Ablation study on adaptive early ending and octree SDF priors on the entire data sequence. We compared changes in the PSNR for different fixed iteration times in Room0 (left) and Room1 (right) of Replica\cite{straub2019replica}. The \textcolor{blue}{blue curve} and \textcolor{Green}{green curve} represent the results with and without octree SDF priors, respectively. The \textcolor{Red}{red star} denotes the average iteration time and PSNR using adaptive early ending.
			% 	}
		% 	\label{fig:iters_ending}
		% 	\vspace{-0.2cm}   
		% \end{figure}
	\begin{table}[t!]
		\centering
		\caption{Ablation study of our design choices}
		\label{tab:ablation}
		\resizebox{\linewidth}{!}{
			\begin{tabular}{c||c||cccc}
				\Xhline{2\arrayrulewidth}
				\shortstack{\\Expanded voxels \\ allocation}  & \shortstack{\\Keyframe \\ startegy}    
				& \shortstack{\\PSNR$\uparrow$\\ $[db]$}          & \shortstack{\\Acc.$\downarrow$\\$[cm]$}   & \shortstack{\\Comp.$\downarrow$\\$[cm]$}   & \shortstack{\\Comp. Ratio$\uparrow$\\$[<5cm\%]$}  \\ \hline
				\XSolid                                       & \Checkmark                                   & 34.25          & 1.16          & 1.26   & 98.52       \\
				\Checkmark                                        & \XSolid                                  & 33.65          & 1.28          & 1.20   & 99.12      \\
				\Checkmark                                        & \Checkmark                                   & \textbf{34.49} & \textbf{1.12} & \textbf{1.13} & \textbf{99.25} \\ \Xhline{2\arrayrulewidth}
			\end{tabular}
		}
		\vspace{-0.3cm}
	\end{table}

	\subsubsection{Adaptive Early Ending} The blue curve in Fig. \ref{fig:iters_ending} shows that increasing the number of iterations leads to higher PSNR\cite{hore2010image} values, but the rate of improvement gradually slows down. As the number of iterations varies when using the adaptive early ending, we calculate the average iteration time and the corresponding PSNR\cite{hore2010image}, represented by the red star. The results demonstrate this strategy adaptively leads to different average iteration times that are close to convergence in various scenarios, which helps to reduce optimization time without compromising accuracy.
	
	% \subsection{Adaptive Early Ending} In \autoref{fig:iters_ending}, we show that adaptive early ending strategy has an average number of iterations that tends to converge for different scenarios. Room0 tends to converge after 6 iterations, while Room1 does so after 7 iterations. The strategy can adaptively select the correct average number of iterations.This can reduce training time for mapping without affecting accuracy. Note that we have chosen the color metric to evaluate the effectiveness of this strategy, as appearance is more difficult to converge.

	\subsection{Real-World SLAM Demonstration}
	\label{subsec:real-world-exp}
	We demonstrate our mapping method with a tracking module, which completes a SLAM system on a handheld device and quadrotors. Specifically, we employ the Realsense L515 as the vision sensor to provide RGB-D images, and a modified VINS-Mono\cite{qin2018vins} incorporating depth constraints as the tracking module to estimate the pose. The handheld device is powered by AGX Orin, and the quadrotor is equipped with Orin NX. All the programs are running onboard. Fig. \ref{fig:real_world_exp} illustrates the results of our real-world experiments. The extracted mesh showcases high-quality reconstruction in both geometry and appearance. We use the handheld device to reconstruct an apartment (Mesh surface: $\approx127m^2$) and use the quadrotors for mapping a part of the fight arena (Mesh surface: $\approx58m^2$). The final mesh is extracted by marching cube\cite{lorensen1987marching} and all optimization was performed within the mapping procedure without any post-processing or additional training time. To the best of our knowledge, our method is the first to achieve high-quality NeRF-based mapping in real-time on edge computers. The specific runtime is evaluated in Sec.\ref{runtime} and more details can be found in the attached video\footnote{https://youtu.be/oR9MlfL86Vw?si=YbemRqVFVLnaZGUi}.

	% \usepackage{multirow}
	% \begin{figure}
		% 		\centering
		% 		\includegraphics[width=\linewidth]{Image/compare_octree_room0_0.pdf}
		% 		\caption{Ablation study on octree SDF priors on one frame. We choose the first frame in Room0 of Replica\cite{straub2019replica} to evaluate how the mapping performance of the corresponding region changes with increasing optimization iterations.}
		% 	\label{fig:sdf_init_one_frame}
		% 	 \vspace{0cm}
		% \end{figure}
	
	\section{Conclusion}
	We propose H$_2$-Mapping, a novel NeRF-based dense mapping system that utilizes hierarchical hybrid representation and can be deployed on edge computers for real-time and high-quality robot mapping. The coarse geometry is represented explicitly using octree SDF priors for fast initialization and convergence, while high-resolution geometry details and texture are encoded implicitly using multiresolution hash encoding in a memory-efficient manner. Furthermore, we propose a coverage-maximizing keyframe selection strategy to improve the reconstruction quality in marginal areas.
	
	Baseline comparisons demonstrate that our method outperforms both mapping quality and time consumption. Besides, ablation studies show that the hierarchical hybrid representation effectively accelerates geometry and texture optimization, and the proposed keyframe selection strategy guarantees reconstruction accuracy even in edge areas. However, currently, our method cannot handle dynamic objects and long-term pose drifting, and further speed-up is required.

	%The plausible solution is to provide an absolute observation, such as GPS or magnetic compass.
	%Alternatively, some of the proposed modules can be supplied in GPS-free environments with exteroceptive sensors to improve the robustness of estimation.
	
	\newlength{\bibitemsep}\setlength{\bibitemsep}{0\baselineskip}
	\newlength{\bibparskip}\setlength{\bibparskip}{0pt}
	\let\oldthebibliography\thebibliography
	\renewcommand\thebibliography[1]{%
		\oldthebibliography{#1}%
		\setlength{\parskip}{\bibitemsep}%
		\setlength{\itemsep}{\bibparskip}%
	}

	\bibliography{RAL2023}

\end{document}